\documentclass[journal]{IEEEtran}
\usepackage{caption}
\usepackage{subcaption}
\usepackage{bm}
\usepackage{float}
\usepackage{booktabs}
\usepackage{graphicx}
\usepackage{amsmath}

\begin{document}

\title{Weak Edge Identification Nets for Ocean Front Detection}

\author{Qingyang Li, Guoqiang Zhong ~\IEEEmembership{Member,~IEEE}, Cui Xie
\thanks{Q. Li, G. Zhong and C. Xie are with the Department of Computer Science and Technology, Ocean University of China, Qingdao, Shandong, 266100 China (e-mail:gqzhong@ouc.edu.cn; 1194094543@qq.com).}
}


\maketitle

\begin{abstract}
The ocean front has an important impact in many areas, it is meaningful to obtain accurate ocean front positioning, therefore, ocean front detection is a very important task. However, the traditional edge detection algorithm does not detect the weak edge information of the ocean front very well. In response to this problem, we collected relevant ocean front gradient images and found relevant experts to calibrate the ocean front data to obtain groundtruth, and proposed a weak edge identification nets(WEIN) for ocean front detection. Whether it is qualitative or quantitative, our methods perform best. The method uses a well-trained deep learning model to accurately extract the ocean front from the ocean front gradient image. The detection network is divided into multiple stages, and the final output is a multi-stage output image fusion. The method uses the stochastic gradient descent and the correlation loss function to obtain a good ocean front image output.
\end{abstract}

\begin{IEEEkeywords}
Ocean front, edge detection, weak edge identification nets, deep learning.
\end{IEEEkeywords}

\IEEEpeerreviewmaketitle

\section{Introduction}

\IEEEPARstart{O}{cean} front, which is generally considered to be a distinct transition between two or more water bodies, which can be described by horizontal gradients of temperature, salinity, density, velocity, color, chlorophyll and other factors. The ocean front generally defined by a ¡°sharp change¡± gradient of some marine environmental parameters, it has an important impact on marine science and atmospheric science. Atmospheric circulation and climatic zones are closely related to large-scale ocean fronts. Due to the dramatic changes in water temperature elements near the ocean front at small and medium scales, the exchange of various weak elements between sea and air is extremely active, so sea storms are easily formed near the ocean front. Large-scale ocean fronts are closely related to the circulation of oceanic water masses. The medium and small-scale ocean fronts are closely related to regional water masses, water systems and circulation. 
\par
The ocean front area is often accompanied by different water bodies carrying nutrients, and plankton is multiplied. It is an important bait area for zooplankton and fish, and is therefore closely related to large fisheries. The spatial and temporal changes of the ocean front have a great impact on the central fishery, fishing season and catch. The marine environment parameters in the ocean front area have changed drastically, and the characteristics of seawater acoustics have changed drastically, which has a great impact on hydroacoustic communication monitoring and submarine activities. Therefore, effective detection of the ocean front is a significant task.
\par
Ocean front as a line with weak edge properties, traditional edge detection methods like Canny operator \cite{1Canny} and Sobel operator \cite{2Sobel}, etc. can not extract the ocean front very well. In recent years, deep learning has performed well in solving many problems such as image object recognition, speech recognition, and natural language processing. Among various types of neural networks, convolutional neural networks(CNNs) are the most intensively studied. In the early days, due to the lack of training data and computing power, it was difficult to train high-performance CNNs, and the rapid improvement of GPU performance made the study of CNNs rapidly. Deep learning theory that solves complex problems with an end-to-end approach is increasingly favored. These methods do not require manual creation of domain-specific knowledge, but rather a qualitative representation from rich data. Deep learning has also been introduced into the fields of earth sciences and remote sensing for data interpretation, analysis and application. Since the deep network has a strong ability to automatically learn the high-level representation of natural images, there is a recent trend of using deep networks to perform ocean front detection. \cite{6Estanislau Lima}, \cite{20Xin Sun.} is to use CNNs to determine whether the ocean front is included in the ocean front gradient image.

\par
The development of neural networks has finally begun to shine in 2012 after decades of troughs. AlexNet's proposal has refreshed ImageNet's record. The use of the nonlinear activation function rectified linear units(ReLU) and dropout method greatly saves the amount of calculation, while ensuring the sparseness of the neural network and alleviating the occurrence of over-fitting problems. Later, VGGNet used more network layers to get more information in the image. Summarizing the above deep learning models and common techniques, we propose a weak edge identification nets (WEIN) for ocean front detection, WEIN contains rich CNNs for ocean front feature extraction, and each convolution layer provides weak edge information to ensure the reliability of the output.
\par
Specifically, we propose a method for detecting ocean front based on weak edge identification. The deep network continuously optimizes the loss function by learning the artificially labeled ocean front information, extracts the weak edge information hidden in the image, and keeps the output close to the true ocean front distribution. The loss function used in this method, in addition to the binary cross entropy loss function commonly used by edge detection algorithms, we also added the Intersection over Union(IoU) loss function for ocean front detection, which is more targeted and effectively improves the accuracy of the detection results. 
\par
Most importantly, although there has been a lot of work on ocean front detection so far, no one has proposed a more scientific quantitative measure of ocean front accuracy. One of the main reasons is that the ocean front has a weak edge properties and it is difficult for people to get an accurate groundtruth is used as an alignment; in addition, the use of edge detection for ocean front detection also requires a large number of manually labeled images as groundtruth to improve the efficiency of neural network training; however, the existing labeled ocean front data is rare. In order to solve this problem, we found experts in the field of ocean front to calibrate the ocean front gradient image within one year, found the ocean front hidden in the image, and made a general ocean front data set according to the needs of the network. Solved the problem of data set blanks in the field of ocean front detection. In the quantitative measurement of accuracy, we use F1 score, Intersection over Union(IoU) and Structural Similarity Index(SSIM) to measure, operate at the pixel level, the experimental part will describe in detail.
\begin{figure}[H]
\centering
\includegraphics[scale=0.4,width=6cm, height=6cm]{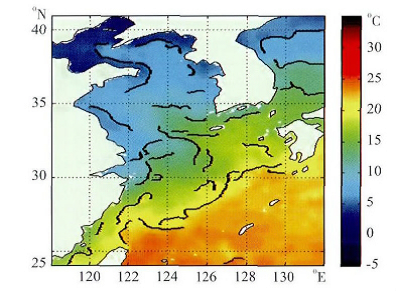}
\caption{A dual-threshold ocean front detection method}
\label{fig:label}
\end{figure}
\section{Related Work}
\subsection{Ocean Front Detection}
Since the Ocean Fronts are important for atmospheric science, fisheries, and military applications, many studies have been done to detect and recognize ocean fronts, among them, the algorithm based on gradient and edge detection has been widely used. For example, Hopkins et al. \cite{3Hopkins} proposed a new statistical modeling approach for detecting and monitoring ocean fronts from Advanced Very High Resolution Radiometer (AVHRR) SST satellite images that builds on a previous front following algorithm, this method finds the ocean front by detecting sudden changes in sea surface parameters. Zhang et al. \cite{4Wei Zhang} proposed a canny-based edge detection algorithm and added binary reconstruction to obtain more continuous ocean front detection results, however, this method requires a complex algorithm to reprocess the output ocean front image to get a good output. Miller \cite{5P. Miller} uses a composite front map approach to combined the location, intensity, and persistence into a single map over a few days, enabling intuitive interpretation of mesoscale structures, But the acquisition of composite front maps is not easy. [15] uses two thresholds $T_1$ and $T_2$ ($T_1$ $\leq$ $T_2$) to obtain two threshold edge connection maps $N_1[i,j]$ and $N_2[i,j]$. Since $N_2[i,j]$ is obtained using a high threshold, it contains few false edges, but often has a breakpoint. The double threshold method is to connect the edges into contours in $N_2[i,j]$. When the breakpoints of the contours are reached, the edges in the eight connected neighborhoods of $N_1$ can be connected to the edges. The algorithm continually collects edges in $N_1[i,j]$ until no edge points can be found that can be joined to the contour, so that an outline is complete, the results are shown in Fig. 1.\par

\cite{20Xin Sun.} proposed a multi-scale ocean front detection and fine-grained location method, which uses AlexNet to judge whether the image contains ocean front, and simultaneously scales the image to be calibrated to different scales, and combines the classification results of the classifier to obtain the ocean front prediction image. \cite{21Mimi Xu.} studied the effect of the ocean front on the season and the atmosphere, and the precipitation structure near the front will also be affected.
Because the ocean front plays an important role in various fields, some scholars have named the ocean front of the Yellow Sea and the
Bohai Sea, as shown in Fig. 2.
\begin{figure}[ht]
    \centering
        \begin{subfigure}[t]{1in}
        {
            \includegraphics[scale=0.3]{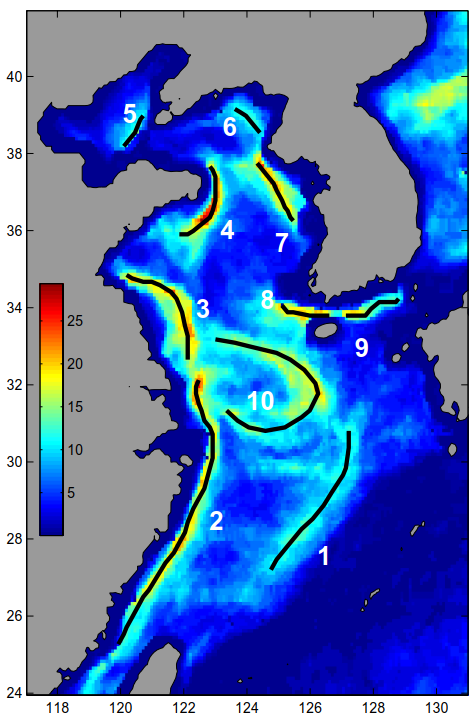}
            \label{fig:label}
        }
        \end{subfigure}
        \quad\quad\quad\quad
         \begin{subfigure}[t]{1in}
        {
            \centering
            \includegraphics[scale=0.5]{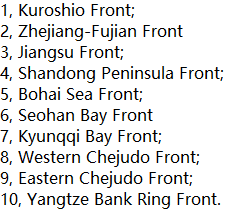}
            \label{fig:label}
        }
        \end{subfigure}
        \caption{Ocean front of the Yellow Sea and the Bohai Sea}
        \label{fig:label}
\end{figure}
\par
\subsection{Edge Detection}
Because the ocean front has weak edge characteristics, we apply edge detection to ocean front detection and hope to acquire more weak edge information. Here we focus on a few representative works. Specifically, the works can be divided into the following categories.
\par
1) \emph{Early Pioneering Methods} like canny detector \cite{1Canny}, sobel detector \cite{2Sobel}, Laplace Operator and LOG Operator, etc. which can effectively detect strong edge information. The Canny edge detection operator has a good detection effect on small-area targets, and the edge discontinuity rarely occurs. But at the same time it also zoomed in on some small areas of the target, and even some of the inconspicuous borders on the background were strengthened. The Sobel detector performs edge detection based on the phenomenon that the pixel points are up and down, and the left and right adjacent gray-scale weighted differences reach extreme values at the edges. It has a smoothing effect on noise, provides more accurate edge information, and the edge positioning accuracy is not high enough. It is a commonly used edge detection algorithm when the accuracy requirement is not very high. The Laplace operator is a second-order differential operator, it is sensitive to gray-scale mutations, has high positioning accuracy, is sensitive to noise, and cannot obtain information such as edge direction, the LOG operator is also a second-order differential operator, which is an improvement on the Laplace operator. Since the Gaussian filtering operation is performed on the image first, a better edge detection effect is obtained than the Laplace operator. Similar methods such as \cite{22D.R.}, \cite{23P.A.} do edge detection by extracting the manually designed features, or local clues of gradients, colors, brightness, and textures. These methods work well for simple computer vision tasks, but are not satisfactory for ocean front detection tasks with weak edge properties because the ocean front does not have such strong edge features.
\par
2) \emph{Learning-Based Methods} that rely on human design are like Multi-scale \cite{9X. Ren}, Structured Edges \cite{10P. Dollar}, etc. Among them, Multi-scales use a local boundary cues including contrast,  localization and relative contrast to train a classifier to integrate them across scales. This approach combines the advantages of large-scale detection (robust but poor  localization) and small-scale detection (retaining details but being sensitive to clutter). This type of method yields better edge detection results, but at the same time it takes more manpower to obtain reliable features, and the results strongly depend on the characteristics of the manual design.
\par
3)\emph{ Convolutional Neural Networks} for edge detection are now also popular, For example, Deep-contour \cite{11W. Shen}, Holistically-Nested Edge Detection \cite{12S. Xie}, Richer Convolutional Features for Edge Detection \cite{13Yun Liu} have achieved good results. Deep-contour uses a CNNs to obtain the deep features of the image, and then enters the visualized feature map into a structured random forest to obtain a contour image. HED is a successful case of edge detection using an end-to-end network. It extracts the feature maps of multiple stages of CNNs, and combines the learning features of different convolutional layers to obtain image edge images without inputting other manually designed features. Image processing technology has become increasingly mature, and its application in many fields has achieved success. Edge detection is the most basic but difficult problem in image processing. Although there are many detection methods at present, they are not well applied to the ocean front detection task. A good ocean front detection method requires the ability to detect an effective ocean front edge, and on the other hand requires a strong anti-noise capability, and expects its calculation to be as small as possible, and our proposed WEIN meets these requirements.
\par
In recent years, the development of deep learning has greatly accelerated the process of image processing, some people have begun to use the deep learning method for ocean front detection. Previous work \cite{6Estanislau Lima} using CNNs can detect the presence of ocean fronts in remote sensing image patches, but still not doing well in the locate of the ocean front. We believe that the end-to-end extraction network can extract the edge information of the ocean frontier more effectively. Therefore, a new end-to-end training method WEIN based on deep learning edge detection is proposed, which has achieved good output.

\section{Data Set Description}
In the experiments, we use our own ocean front data set to evaluate WEIN. The data set is derived from satellite remote sensing images, and after special treatment, the ocean front gradient image is obtained.

\begin{figure}[ht]
\centering
\includegraphics[scale=1.3]{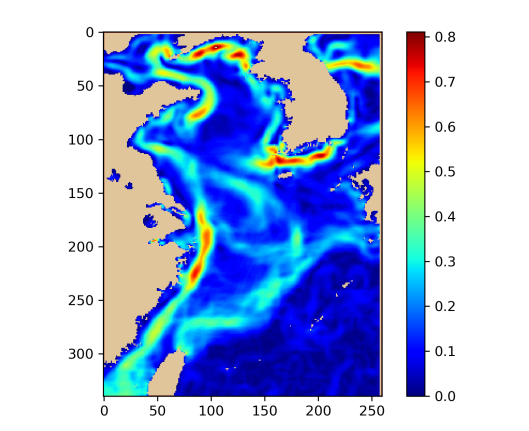}
\caption{Ocean front gradient image example}
\label{fig:label}
\end{figure}

The acquisition of groundtruth is a more important job, we found some experts in the field to obtain calibration. Since the ocean front is constantly occurring and dying with the seasons, the shape of the front is constantly changing at different times, so the experts mark the ocean front according to the gradient images of different periods. The calibration uses the ocean front gradient map shown in Fig. 3. After expert discussion, it is decided to select the vicinity of 0.4 of the right colorimetric rod as the threshold, that is, the area larger than 0.4 in the image can be used as the ocean front. In order to make the results accurate, the calibration of each ocean front gradient image takes the average result of several experts, accurate labeling results are good for improving the performance of neural networks, an example of a calibration image is shown in Fig. 4.

\begin{figure*}[ht]
\centering
\includegraphics[scale=1.2]{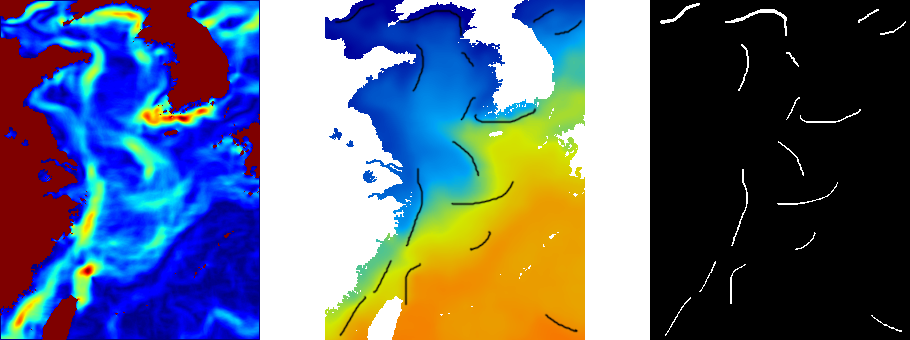}
\caption{From left to right, the ocean front gradient image for training, artificially annotated image, and extracted groundtruth.}
\label{fig:label}
\end{figure*}
\par
Our data set includes ocean front gradient images for the year of 2014. We observed that the ocean front gradient images of each year are not very different, so it is not necessary to collect the ocean front gradient images data for many years, but use the 2014 images to approximate the images of other years, get several years of training data. For the 2014 ocean front gradient images, we randomly select five images as the test set in each month's ocean front gradient images, and the remaining images are used as training set, that is, the training set has 305 images and the test set has 60 images.

\section{Method}

\subsection{The Theoretical Basis of CNNs}

With the development of deep learning and image processing in recent years, CNNs have gradually become a commonly used and efficient tool, and are widely used in object recognition, image classification, segmentation and other visual tasks. This is because deep CNNs with more hidden layers have more complex network structures. Compared with traditional machine learning methods, it has more powerful feature learning and feature expression ability to extract the internal and bottom features of images, avoiding complex prior constraints. In the past, the extraction of ocean front was very difficult because the feature extraction and expression of traditional methods was not strong enough, CNNs solved this problem, so we applied it to this field. Next, we will systematically introduce how CNNs learn the weak edge features of the ocean front.\par
\emph{1) Convolution Layer: }The convolutional layer is the basic unit of CNNs, the feature extraction of the ocean front is inseparable from the contribution of each convolution layer. In the convolutional layer, the feature map of the upper layer is convolved by a learnable convolution kernel, and then an output function map is obtained through an activation function. Each output feature map can be combined to convolve multiple feature map values \cite{7Bouvrie J}.
\begin{equation}
\bm{x}^l_{j} = f(\bm{u}^l_{j})
\end{equation}
\begin{equation}
\bm{u}^l_{j} = \sum_{i \in M_j}(\bm{x}^{l-1}_{i} * \bm{k}^l_{ij} + \bm{b}^l_j)
\end{equation}
\par
Among them, $u^l_j$ is called the net activation of the $j_{th}$ channel of the convolutional layer $l$, which is obtained by convolution summation and offset operation on the previous layer output characteristic map $x^{l-1}_i$, and $x^l_j$ is the first of the convolution layer $l$ the output of $j$ channels. $f(\cdot)$ is called the activation function, usually $sigmoid$, $tanh$ and $ReLU$ can be used. $M_j$ denotes the input feature map subset used to calculate $u^l_j$, $k^l_{ij}$ is the convolution kernel matrix, and $b^l_j$ is the offset of the convolutional feature map. For an output feature map $x^l_j$, each input feature map $x^{l-1}_i$ corresponds to the convolution the kernel $k^l_{ij}$ may be different, $*$ is a convolutional symbol.
\par
\emph{2) Back Propagation Algorithm: }There are two basic modes of operation for neural networks: forward propagation and back propagation. Forward propagation is the process by which an input signal passes through one or more network layers in the previous section and then outputs at the output layer. Back propagation algorithm is a commonly used method in supervised learning of neural networks, Its goal is to estimate network parameters based on training samples and expected output. Taking our WEIN as an example, the main task of CNNs is to learn the convolution kernel parameter $k$, the downsampling layer network weight $\beta$ and offset parameters of each layer $b$, the essence of the back propagation algorithm is to calculate the effective error of each network layer. and thus derive a learning rule for network parameters, making the WEIN output image closer to the real ocean front distribution [8]. The improved binary cross entropy loss function is used in our method, and we take this loss function as an example to introduce the idea of back propagation algorithm, consider the total training error of a two-class problem, defined as the difference between the expected output value of the output and the actual output value.
\begin{equation}
E(\bm{\beta},\bm{k},\bm{b}) = -\frac{1}{N}\sum^N_{i=1}[y_ilog\hat{y_i} + (1-y_i)log(1-\hat{y_i})]
\end{equation}
\par
Where $y_i$ is the category label groundtruth of the $i_{th}$ sample, and $\hat{y_i}$ is the category label of the $i_{th}$ sample predicted output through the forward propagation network. In the process of back propagation, the neural network obtains the difference between the predicted image and the real ocean front distribution through the loss function, propagating the gradient from the back to the front and updating the parameters to reduce the distance between the two, when the predicted image can obtain accurate and reliable ocean front output, the learning of weak edge information is realized.
\begin{figure*}[ht]
\centering
\includegraphics[scale=0.625]{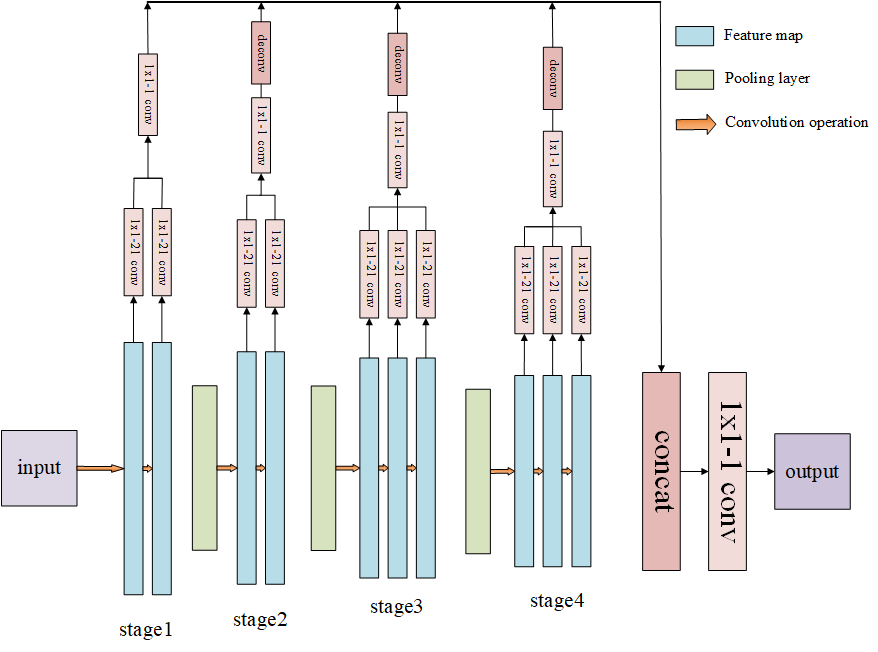}
\caption{The overall architecture of the proposed ocean front edge detection network consists of four stages, with the first two stages containing two convolutional layers and the last two stages containing three convolutional layers. Each convolutional layer is connected to a 1 $\times$ 1 convolutional layer with a channel depth of 21, and the obtained output is eltwise with the output of the same stage. After being sent to 1 $\times$ 1 - 1 conv, the feature map is obtained through the upsampling layer. Among them, the loss of each step is generated after the upsampling.}
\label{fig:label}
\end{figure*}

\subsection{Proposed Ocean Front Detection Framework}
For the problem that most methods can't detect weak edge information, the proposed framework can effectively obtain the weak edge information of the ocean front. We created a deep convolutional network with four stages to get the feature map of the ocean fronts for each stage and fuse them to get the final ocean front boundary results.\par
In addition to using the commonly used edge detection cross entropy loss function, we also designed the IoU loss function and added a variable loss for ocean front detection. By comparing the calibration data and the detection results from the pixel level, the difference between the two is minimized. The experimental results show that our method is effective.

\subsection{Network Architecture}
Inspired by the deep edge detection network, we propose an edge detection framework for the ocean front, as shown in Fig. 5. The network uses CNNs to extract weak edge features, and joins pooling layers to reduce the image size and parameters. The specific structure is as follows: \par
(a) In order to take full advantage of the advantages of CNNs and avoid excessive parameters of the full connection layer, our method only uses CNNs instead of the fully connected layer, which can greatly reduce the storage overhead and computational effort. The entire network consists of four stages. For the first two stages, each contains two \emph{conv} layers; for the last two stages, each contains three \emph{conv} layers. The transition between the stages is done through the pooling layer. \par
\begin{figure*}[ht]
\centering
\includegraphics[scale=0.4,width=18cm, height=4.7cm]{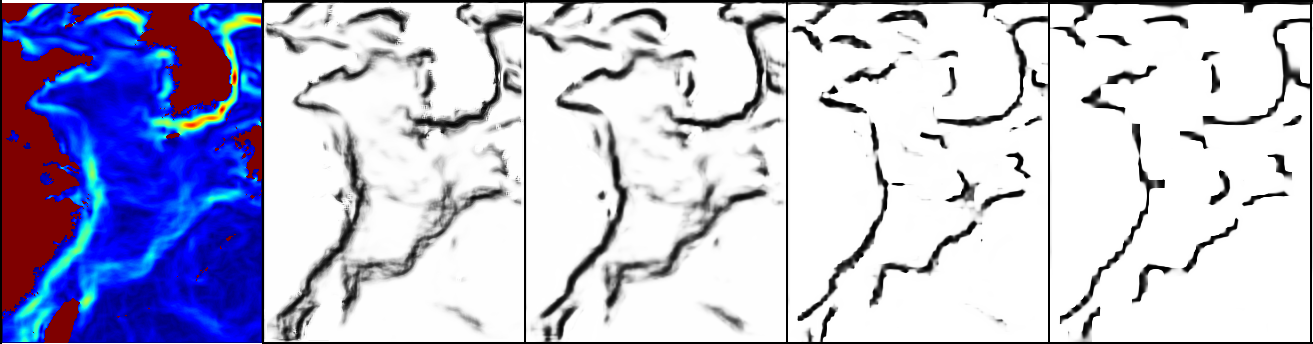}
\caption{The output of each stage of our method, the leftmost is the original image to be predicted, and the second image to the fifth image represent the output of stage 1, 2, 3, 4 respectively.}
\label{fig:label}
\end{figure*}
\par
(b) For each \emph{conv} layer, we are followed by a side output layer, and the side output layer is calculated using a 1 $\times$ 1 \emph{conv} kernel with a channel depth of 21. The result of each stage side output layer uses the \emph{eltwise} layer to attain hybrid features. The role of the side output layer is to obtain a more richer edge information of the ocean front, so that each \emph{conv} layer is fully functional.\par
(c) There is a 1 $\times$ 1 - 1 \emph{conv} layer after each \emph{eltwise} layer. In the 2,3,4 stage, a \emph{deconv} layer is used to up-sample this feature map, the first stage does not require a \emph{deconv} layer because the feature map obtained in this stage is identical to the output image size.\par
(d) The feature map obtained by each upsampling layer will get a loss. At the end, WEIN will get the fusion loss of four stages.\par
With the above design, WEIN can collect all the ocean front weak edge information of the \emph{conv} layer to achieve more accurate ocean front detection. Since the receptive field sizes of \emph{conv} layers in the network are different, as shown in Table I, multi-scale edge information can be learned, which contributes to the extraction of weak edge detection. Fig. 6. shows the intermediate results of each stage. From left to right, the \emph{conv} layers learn more abundant weak edge information as the receptive field increases, and gradually becomes more sensitive to the ocean front edge response, ignoring the noise impact. This has met our expectations.
\begin{table}[h] 
\caption{The receptive field and stride size in ocean front edge detection network.} 
\begin{tabular}{p{1cm}|p{1cm}|p{1cm}|p{1cm}|p{1cm}|p{1cm}} 
\hline
\hline
layer & conv1\_1 & conv1\_2 & pool1 & conv2\_1 & conv2\_2\\
\hline 
rf size & 3 & 5 & 6 & 10 & 14\\
\hline
stride & 1 & 1 & 2 & 2 & 2\\
\hline
\hline
\end{tabular}
\begin{tabular}{p{1cm}|p{1cm}|p{1cm}|p{1cm}|p{1cm}|p{1cm}} 
\hline
\hline
layer & pool2 & conv3\_1 & conv3\_2 & conv3\_3 & pool3\\
\hline 
rf size & 16 & 24 & 32 & 40 & 44\\
\hline
stride & 4 & 4 & 4 & 4 & 8\\
\hline
\hline
\end{tabular}
\begin{tabular}{p{1cm}|p{1cm}|p{1cm}|p{1cm}|p{1cm}|p{1cm}} 
\hline
\hline
layer & conv4\_1 & conv4\_2 & conv4\_3 & &\\
\hline 
rf size & 60 & 76 & 92 & &\\
\hline
stride & 8 & 8 & 8 & &\\
\hline
\hline
\end{tabular}
\end{table}
\subsection{Loss Function}
There are two kinds of loss functions. The first one is the binary cross entropy loss function commonly used in edge detection. It can learn the basic edge information, but the front line detected by the cross entropy loss function is very thick. We added the IoU loss function, which can effectively optimize the ocean front line of the output, Next we will detail these two loss functions.
\par
\emph{1)  Binary Cross Entropy Loss Function: }In the edge detection task, we assign a probability value to each pixel of the image, which represents the probability that the pixel is the edge point. When classifying, we set a threshold \emph{p} between 0 and 1. When the probability value of a pixel is greater than \emph{p}, the pixel is classified as an edge pixel. Conversely, when the probability of the pixel is less than \emph{p} This point is classified as a non-edge pixel. However, in an image, the number of edge points may only be a small fraction of the total number of pixels, while the remaining pixels are non-edge points. Therefore, when calculating the loss, the optimizer tends to classify all pixels as non-edge points, so the total loss will be minimal, but it is obvious that the result will be poor. Therefore, we have to assign different cost to edge pixels and non-edge pixels. When edge pixels are misclassified, there will be a greater cost loss. This is like a binary classification problem in which the number of positive samples is much smaller than the number of negative samples. In order to balance the total cost of the positive and negative samples, a larger loss is obtained when the positive samples are classified incorrectly. We obtain a loss function for ocean front detection by improving the binary cross entropy loss:

\begin{equation}
loss_{bce} = \begin{cases}
r(1-\beta)\cdot log(1-S(\bm{X}_i;\bm{W})),& y_i \in U_{non-edge}\\
\beta\cdot log(S(\bm{X}_i;\bm{W})),& y_i \in U_{edge}
\end{cases}
\end{equation}
where:
\begin{equation}
\beta = \frac{\mid Y_- \mid}{\mid Y_+ \mid + \mid Y_- \mid}
\end{equation}
$\mid Y_+ \mid$ represents the number of edge pixel points and $\mid Y_- \mid$ represents the number of non-edge pixel points, $\beta$ represents the cost of the edge pixel point classification error, and its value is obtained by dividing the number of non-edge pixel points by the total number of pixel points. Because the number of edge pixels is small, there is a greater loss when classified incorrectly, so the value of $\beta$ will be larger. \emph{r$\cdot$(1-$\beta$)} is the cost of non-edge pixel points, where \emph{r} is the hyper-parameter of the balanced edge pixel and non-edge pixel, because the ocean front line detected by the deep neural network tends to be thicker, we set \emph{r} to with large values, deep networks are more cautious when classifying pixels as ocean fronts, resulting in a finer ocean front. $X_i$ and \emph{W} are the input distribution and weight matrix of the $i_{th}$ pixel, respectively, and $y_i$ is the value of groundtruth, which consists of two sets $U_{edge}$ and $U_{non-edge}$, which represent the set of edge pixels and the set of non-edge pixels. $S(\cdot)$ is a standard sigmoid function. The total loss of the model is calculated from the side output feature map of the four stages and the final fusion feature map.
\par
\emph{2) IoU Loss Function: }
The ocean front line learned by the deep neural network using only the cross entropy loss function will be coarser, because the image features of different regions of the ocean front gradient image may be similar, and the area around the ocean front line will be misclassified by the neural network. In order to correct this error and get a more detailed ocean front output, we propose an IoU loss function for ocean front edge detection, which is mainly inspired by \cite{14Borui Jiang}. By observing the results of the conventional deep learning edge detection algorithm \cite{12S. Xie} ocean front detection, we found that the number of ocean edge edge pixels detected by the model is much larger than the actual ocean front edge pixels. We want to divide the intersection of the edge pixel of the predicted image and the edge pixel of the groundtruth by the union of the two as the IoU ratio, and hope that the closer it is to \emph{1}, the better. At the same time, since the ocean front pixel of the predicted image almost covers the oceanfront pixel of the groundtruth, we use the approximation of the total number of edge pixels of the predicted image to represent the union, the value of intersection divided by union is fitted to \emph{1}, so that the values of the two are closer to each other, and the refined ocean front output is obtained.
General:
\begin{equation}
IoU = \frac{PredictionResult \cap Groundtruth}{PredictionResult \cup Groundtruth}
\end{equation}
As mentioned above, since the groundtruth of the ocean front is very thin and the front line of the prediction result is very thick, we use the total number of edge pixels of the groundtruth as the intersection, and the total number of edge pixels of the predicted image as the union:
\begin{equation}
IoU = \frac{Groundtruth}{PredictionResult}
\end{equation}
The value of the IoU is a number greater than 0 and less than 1, We match this value to 1 to reduce the number of predicted edge pixels, resulting in a predicted image that is closer to the true ocean front distribution.
Therefore, the loss function of each pixel consists of the following parts, which are the binary cross entropy $loss_{bce}$ commonly used for edge detection tasks, and the IoU loss and a variable loss(In the experiment, we tried a variety of loss functions to fit the predicted image distribution and groundtruth distribution, and finally chose the SmoothL1 loss, but other types of losses can also be used or not):
\begin{equation}
l = loss_{bce} + loss_{iou} + loss_*
\end{equation}
From the image-level, for an image with a number of pixels of \emph{I}, the total loss is the distance between the feature map of each stage or fusion prediction and the groundtruth:
\begin{equation}
L = \sum_{i=1}^I\Big(\sum_{m=1}^Ml(\bm{X}_i^{(m)}; \bm{W}) + l(\bm{X}_i^{(fuse)}; \bm{W})\Big)
\end{equation}
Here, $X_i$ and \emph{W} have the same meaning as in Equation 4. \emph{M} represents the number of stages, here is equal to 4.\par
During testing, given an image, we obtain the edge prediction map through the side output layer and the weighted fusion layer. We further aggregate the edge prediction map to obtain the final unified output:
\begin{equation}
\bm{\widehat{Y}} = Average(\bm{\widehat{Y}}_{fuse}, \bm{\widehat{Y}}_{side}^{(1)}, ..., \bm{\widehat{Y}}_{side}^{(M)})
\end{equation}
\section{Experments}

\subsection{Data Set Introduction and Model Matching}
The temperature gradient is the physical quantity that describes the direction in which the temperature changes most rapidly within a particular regional environment, and at what rate. Obviously, the temperature gradient will change over time, and the day-to-day or seasonal hot and cold transition will cause the temperature gradient to change. Changes in the ocean temperature gradient directly affect the change of the ocean front. Therefore, we use the temperature-dependent ocean front gradient images for each day of the year as the research object. Specifically, firstly, the remote sensing images collected by the satellite is processed to obtain a ocean front gradient images suitable for CNNs operation; then the appropriate tools and indicators are selected to mark the groundtruth; finally, the data set is sent to the ocean front detection network, and fine-tuning to get the results.\par
Traditional edge detection methods such as canny operator, sobel operator can also do some simple edge detection work, but these methods are affected by strong edge information, can not detect the weakly edge information of the ocean front, and often detect the edge of the land, the accuracy of the results is largely destroyed. The advantage of the deep learning method is that the weak edge information in the image can be obtained by learning the feature. Our method has unique advantages in the weak edge identification task. \cite{12S. Xie} only merged the output feature map of the last convolution layer of each stage, ignoring the role of other convolutional layers, in order to make full use of all useful information obtained by the convolutional layer, we collect the feature information of each convolution layer \cite{13Yun Liu}. In addition, RCF \cite{13Yun Liu} with 5 stages changes stride of pool4 layer to 1, and uses the atrous algorithm to fill the holes. Even so, the feature map of the stage5 output contains very little weak edge information, We do not think it makes much sense to do this, so pool4 and stage5 are directly removed to fuse the more reliable feature maps. Our approach not only combines the advantages of these two methods, but also avoids their drawbacks, showing excellent performance in the ocean edge weak edge detection task.

\par
\subsection{Evaluation Metrics} In recent years, with the maturity of edge detection technology, traditional edge detection methods and deep learning edge detection methods have achieved certain results, and some scholars have contributed to the detection of ocean front \cite{18Di Xiao.}, \cite{19Biao Chen.}, but compared with the conventional Computer vision tasks such as object detection, semantic segmentation, etc., the work on the ocean front is still so small that there is no universal metric to judge the quality of the ocean front detection results. This paper modifies three commonly used metrics in the field of computer vision to measure the results of ocean front detection. Specifically, the three methods are F1 score, Intersecton over Union(IoU) and Structural Similarity Index(SSIM). The three methods will be introduced separately below.\par

\emph{(a) F1 Score as a Metrics:} Error rates and accuracy are common metrics but are not applicable to ocean edge detection tasks. The ocean front edge detection task can be seen as a two-class task for image pixels. We must not only determine how many real ocean front pixels have been successfully detected, but also know which pixels in the predicted image are misclassified. The recall and precision are the performance metrics that are more suitable for this situation. For the ocean front detection problem, we divide the sample into true positive(TP), false positives(FP), true negative(TN), and false negative(FN) according to the combination of its groundtruth and prediction results. Among them, the precision is:
\begin{equation}
P = \frac{TP}{TP+FP}
\end{equation}
The recall is:
\begin{equation}
R = \frac{TP}{TP+FN}
\end{equation}
Here, the precision and recall are contradictory. When the precision is high, the recall is often low; when the recall is high, the precision is often low. The F1 score is based on the harmonic mean of the precision and recall , defined as:
\begin{equation}
F1 = \frac{2 \times P \times R}{P+R}
\end{equation}
Since the harmonic mean is less affected by extreme values, the F1 score is more suitable for evaluating the classification of unbalanced data. Therefore, we use F1 score as a measure of ocean edge detection.\par
In particular, since we use a thin line with a width of one pixel to get the groundtruth, and the real ocean front is slightly wider, we used a method to avoid this error when calculating the true positive. We put the groundtruth and its neighborhood is a true positive. Specifically, we take two lines closest to each other on the left and right sides of the calibrated front line, we also use the two closest lines as a true positive. The number of ocean front pixel points obtained by this method is closer to the true distribution.\par
\emph{(b) IoU as a Metrics: }The ocean front differs from the edges or boundaries in a conventional image. It is more difficult to detect as a weak edge than a conventional edge or boundary, and since the front line is a gradual process, the true ocean front distribution tends to be thicker on the image. It is an edge with a width. According to the nature of the ocean front, and inspired by the object detection task, we decided to measure the quality of the test results by querying the intersection of the test results and the groundtruth. and using the IoU metric for weak edge detection, which compares the difference between the predicted distribution and the groundtruth to more accurately measure the weak edge detection task. Although our mission is different from object detection, the metric is somewhat compatible with our mission. The ratio of the ocean front distribution to the true ocean front distribution in the predicted image can effectively characterize the performance of the ocean front classifier. Specifically, we compare the edge points of the predicted image with the edge points of the groundtruth by position. The edge point of the predicted image and the edge point of the groundtruth are combined as a union. Then, we intersect the predicted edge points of the image with the groundtruth edge points and their neighborhoods. The calculation method of the groundtruth edge point neighborhood is similar to the true position of the F1 score, the two nearest neighbors on the left and right sides of the groundtruth are also used as intersections. The purpose of this is to eliminate the errors caused by manual labeling.\par
\emph{(c) SSIM as a Metrics:} Structural similarity index, SSIM index is a metric to measure the similarity between two images. Structural similarity is more in line with the human eye's judgment on image quality in the measurement of image quality. The basic concept of structural similarity is that images are highly structured \cite{24Zhou Wang}, and there is a strong correlation between adjacent pixels in the image. When measuring the results of ocean front detection, it is an important part of structural discrimination.
\par
Specifically, given two signals $\bm{x}$ and $\bm{y}$ (in our mission, these two signals are predicted images and groundtruth, respectively), the structural similarity of the two is defined as:
\begin{equation}
SSIM(\bm{x},\bm{y}) = [l(\bm{x},\bm{y})]^{\alpha}[c(\bm{x},\bm{y})]^{\beta}[s(\bm{x},\bm{y})]^{\gamma}
\end{equation}
Where:
\begin{equation}
l(\bm{x},\bm{y}) = \frac{2\mu_x\mu_y + C_1}{\mu^2_x+\mu^2_y+C_1}
\end{equation}

\begin{equation}
c(\bm{x},\bm{y}) = \frac{2\sigma_x\sigma_y + C_2}{\sigma^2_x+\sigma^2_y+C_2}
\end{equation}

\begin{equation}
s(\bm{x},\bm{y}) = \frac{\sigma_{xy} + C_3}{\sigma_x\sigma_y+C_3}
\end{equation}
Among them, $l(\bm{x},\bm{y})$ compares the lightness of $\bm{x}$ and $\bm{y}$, $c(\bm{x},\bm{y})$ compares the contrast of $\bm{x}$ and $\bm{y}$, and $s(\bm{x},\bm{y})$ compares the structure of $\bm{x}$ and $\bm{y}$, $\alpha > 0$, $\beta >0$, $\gamma > 0$, is the parameter to adjust the relative importance of $l(\bm{x},\bm{y})$, $c(\bm{x},\bm{y})$, $s(\bm{x},\bm{y})$, $\mu_x$ and $\mu_y$, $\sigma_x$ and $\sigma_y$ are the mean and standard deviation of $\bm{x}$ and $\bm{y}$, respectively, $\sigma_{xy} $ is the covariance of $\bm{x}$ and $\bm{y}$, $C_1$, $C_2$, $C_3$ are constants to maintain the stability of $l(\bm{x},\bm{y})$, $c(\bm{x},\bm{y})$, $s(\bm{x},\bm{y})$. The larger the value of the structural similarity index, the higher the similarity between the two signals. When the two signals measured are identical, the value of the structural similarity index is 1.\par
In our ocean front detection mission, all ocean front predicted images from the deep network are used as the first input signal, and groundtruth is used as the second input signal, using the SSIM method to measure the similarity between the two. Compared with the first two methods, the results obtained by this method are more consistent with the visual observations of people.
\subsection{Parameter Setting}
Unlike fine-tuning CNN for image classification, adjusting the convolutional network for ocean front edge detection requires special attention. Differences in data distribution, groundtruth distribution, and loss functions all contribute to network convergence difficulties, even when initializing pre-training models. We implemented our network using the publicly available pytorch, which is simple and efficient. BSDS500 \cite{17P.} is a widely used dataset in edge detection. An improved VGG16 model pre-trained on BSDS500 is used to initialize our network. The pre-training model can assign a probability value to each pixel, which can better describe the strong edge information, but predicting the weak edge pixels of the ocean front is not advantageous, so it is necessary to set appropriate hyper-parameters. We use stochastic gradient descent (SGD) for optimization. The important hyperparameters are selected as follows: the learning rate is set to 1e-6, the learning rate decay \emph{Gamma} is set to 0.1, the momentum is 0.9 and the weight deacy is 0.0002. The hyper-parameter \emph{r} needs to be set according to the dataset. On some datasets, setting this value to 1.1 can achieve better results \cite{13Yun Liu}, but for the ocean front dataset we set it to 1.9, which helps prevent overfitting. All experiments in this paper were done using the NIVADA GeForce GTX 1080 Ti.

\subsection{Experimental Results and Comparisons}
This article aims to use the deep convolutional network to learn the weak edge characteristics of the ocean front and accurately extract the ocean front from a given gradient image. Unlike \cite{20Xin Sun.}, our method uses an end-to-end model, which does not require a selection algorithm to extract features from the image. Instead, the model learns to obtain weak edge information without human intervention, which greatly improves efficiency. In addition, when dealing with the problem of weak edge detection, the advantage of deep learning is clearly manifested. The groundtruth obtained by a priori calibration of the ocean front can make the network effectively learn the weak edge information. We can visually see from the detection image, as shown in Fig. 7, the output obtained by the sobel operator is ambiguous, and the boundary between the ocean and the land is also incorrectly marked. Overall, it is far less excellent than the output of our method.

\par
Although Ocean Front has important influences on various fields, due to the scarcity of data, there is no unified quantitative measurement index in this field. To solve this problem, we use the collected ocean front dataset to measure the ocean front detection results in three ways, namely F1 score, IoU and SSIM. The results of F1 score and IoU are shown in Table II, the results of SSIM are shown in Table III. In the specific experiment, we selected a variety of edge detection methods to detect the ocean front dataset, including traditional edge detection algorithms such as canny operator, sobel operator, and also the best effective deep learning methods such as HED, RCF. Since the traditional method is more inclined to select sharp points with sharp gradients as edge points, such as the junction between land and sea, it does not perform well in the ocean front detection mission, the deep learning method is more advantageous in weak edge detection tasks. \par
\begin{table}[h]
\caption{Measurement of ocean front test results using F1 score and IoU quantification, which includes deep learning methods and traditional edge detection methods.}
\large
\begin{tabular}{p{2.4cm}|p{2.4cm}|p{2.4cm}}
\toprule[1pt]
Methods& F1 socre& IoU\\
\midrule
Canny& 19.36\%& 12.13\%\\
Sobel& 22.53\%& 30.33\%\\
HED& 60.23\%& 72.67\%\\
RCF& 61.68\%& 75.88\%\\
Ours& \textbf{62.36}\%& \textbf{78.56}\%\\
\bottomrule[1pt]
\end{tabular}
\end{table}
\par
In the section on introducing the loss function, we added a variable loss ${loss_*}$ in addition to the two required loss functions. The experiment in Table II sets the $loss_*$ to null because it does not improve the F1 score and IoU. The loss exists to obtain intuitively better output, and SSIM is the closest metric to people's visual observations, so we added the $loss_*$ to improve the SSIM score. In the specific experiment, we set the $loss_*$ to L1 loss, MSE loss and SmoothL1 loss respectively, but experiments show that LI loss and MSE loss will make the total loss too large, so the training result can not converge, so choose the most stable SmoothL1 loss. In this way, we can obtain visually clearer ocean front detection results, and the SSIM score comparison is shown in Table III.
\par
After adding $loss_*$, the output we get is closer to groundtruth, that is, the ocean front is finer. However, the true ocean front is a transition zone with a certain width. As far as the true distribution of the ocean front is concerned, it may be better not to add the loss. In practical applications, the choice of $loss_*$ can be determined as needed. The experimental results of the two training methods are shown in Fig. 8.
\begin{table}[h]
\caption{Measurement of ocean front test results using SSIM quantification, which includes deep learning methods and traditional edge detection methods.}
\large
\begin{tabular}{p{3.6cm}|p{3.6cm}}
  \toprule[1pt]
  Methods & SSIM \\
  \midrule
  Sobel &  24.19\%\\
  Canny & 71.50\% \\
  HED & 83.25\% \\
  RCF & 87.64\% \\
  Ours without $loss_*$ & 88.03\% \\
  Ours with $loss_*$ & \textbf{92.06\%} \\
  \toprule[1pt]
\end{tabular}
\end{table}

\begin{figure*}[ht]
  \centering
	\begin{subfigure}[t]{1in}
		\centering
		\includegraphics[scale=0.6]{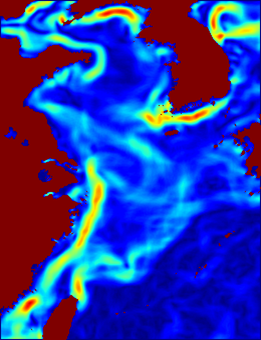}
		\caption{}\label{fig:1a}		
	\end{subfigure}
	\quad\quad\quad\quad\quad
	\begin{subfigure}[t]{1in}
		\centering
		\includegraphics[scale=0.6]{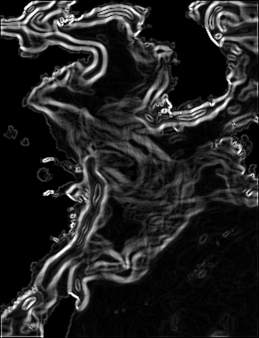}
		\caption{}\label{fig:1b}
	\end{subfigure}
    \quad\quad\quad\quad\quad
    \begin{subfigure}[t]{1in}
		\centering
		\includegraphics[scale=0.6]{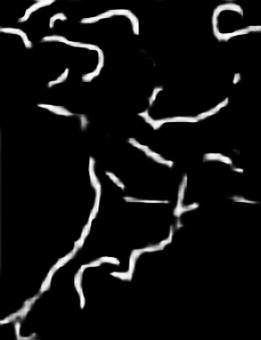}
		\caption{}\label{fig:1c}
	\end{subfigure}
    \quad\quad\quad\quad\quad
    \begin{subfigure}[t]{1in}
		\centering
		\includegraphics[scale=0.3875]{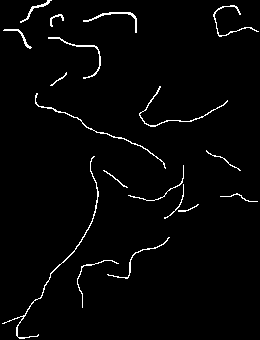}
		\caption{}\label{fig:1d}
	\end{subfigure}
  \caption{The traditional edge detection method and the ocean front detection result of our method. (a) ocean front gradient image. (b) ocean front image detected by sobel operator. (c) ocean front image obtained by our method. (d) groundtruth.}
\label{fig:label}
\end{figure*}

\begin{figure*}[ht]
  \centering
	\begin{subfigure}[t]{1in}
		\centering
		\includegraphics[scale=1.085]{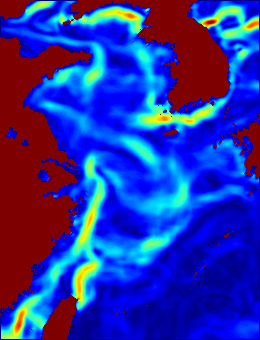}
		\caption{}\label{fig:1a}		
	\end{subfigure}
	\quad\quad\quad\quad\quad
	\begin{subfigure}[t]{1in}
		\centering
		\includegraphics[scale=0.3875]{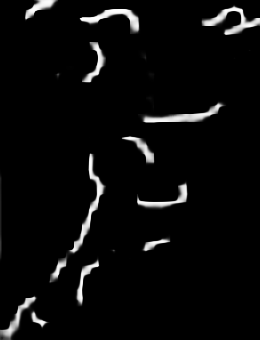}
		\caption{}\label{fig:1b}
	\end{subfigure}
    \quad\quad\quad\quad\quad
    \begin{subfigure}[t]{1in}
		\centering
		\includegraphics[scale=0.3875]{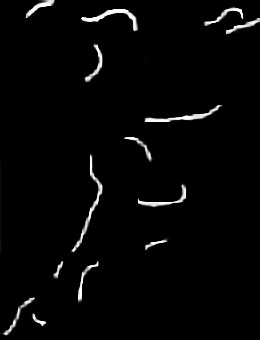}
		\caption{}\label{fig:1c}
	\end{subfigure}
    \quad\quad\quad\quad\quad
    \begin{subfigure}[t]{1in}
		\centering
		\includegraphics[scale=0.3875]{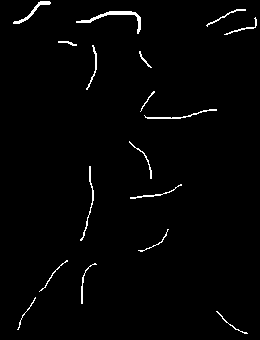}
		\caption{}\label{fig:1d}
	\end{subfigure}
  \caption{The effect of adding $loss_*$ on the test results. (a) ocean front gradient image. (b) ocean front detection image without $loss_*$. (c) ocean front detection image with $loss_*$. (d) groundtruth.}
\label{fig:label}
\end{figure*}
In the hyperparameter settings section we mentioned that balancing the positive and negative examples of hyperparameters \emph{r} plays a crucial role in our mission. \cite{13Yun Liu} limits \emph{r}  to 0.5 to 1.1, which is used for general edge detection tasks. In our mission, \emph{r} plays a more important role. Why? This is determined by the weak edge nature of the ocean front. As a transition zone, the ocean front does not have the same sharp gradient as the edge of the object, nor the fine texture, but a weak edge with a certain width. When deep network learns weak edge information, there will be a tendency to overfit, and the point near the real ocean front is also regarded as the edge of the ocean front, and the number of these points is far more than the number of edge points that make up the ocean front, resulting in a certain degree of deviation in the test results. In order to solve this problem, we set the hyperparameter \emph{r}  to a larger value. When the non-edge points are misclassified, the classifier will get a larger loss, so that it is more cautious in the classification and obtain excellent results, The specific experimental results are shown in Table IV.
\begin{table}[h]
\caption{Our approach takes the results of different hyper-parameters \emph{r}, including F1 score and IoU.}
\large
\begin{tabular}{p{2.7cm}|p{2.25cm}|p{2.25cm}}
\toprule[1pt]
Hyperparameter value& F1 socre& IoU\\
\midrule
\emph{r}=0.5& 61.63\%& 74.74\%\\
\emph{r}=1.1& 62.14\%& 77.04\%\\
\emph{r}=1.9& \textbf{62.36}\%& \textbf{78.56}\%\\
\bottomrule[1pt]
\end{tabular}
\end{table}
\subsection{The Advantage of WEIN}
\par
\emph{1) Advantages of CNNs: }WEIN uses CNNs to extract ocean front features, which maintains a certain local translation invariance in the downsampling layer. The number of parameters that the WEIN needs to train is reduced by the receptive field and weight sharing in the convolutional layer. Each neuron only needs to perceive the local image area to obtain a small part of the ocean front weak edge information. At the higher level, these neurons that experience different local regions can be combined to obtain global ocean front information. Therefore, the number of network connections can be reduced, that is, the number of weight parameters that our network needs to train can be reduced. The weights of neurons are the same, so WEIN can learn in parallel, which is also a major advantage of CNNs relative to fully connected networks.
\par
In short, CNNs have special advantages in weak edge identification compared with general neural networks: a) The network structure can better adapt to the structure of the image; b) Simultaneous feature extraction and classification, so that weak edge features extraction helps edge points feature classification; c) Weight sharing can reduce the training parameters of the network, making our network structure simpler and more adaptable.
\par
\emph{2) The Role of Deep Supervision: }In general, images with weak edge properties are difficult to detect using traditional methods, but deep learning methods can produce good results. What is the reason? It is the role played by deep supervision. Specifically, we use the deep learning model, the entire network is the path-connected and output layer parameters can be back-propagated through the weighted fusion layer to update the propagation path through the weighted-fusion layer, each layer plays an important role. In the back-propagation process, the distance between the output and the groundtruth constitutes the loss of the network. Our goal is to minimize the distance between the two, because groundtruth contains rich weak edge information, which comes from Marking people's understanding of the ocean front, when our model learned this information, it realized the deep network learning of weak edge information, which is the advantage of deep supervision.

\section{Conclusion}
In this paper, we have proposed an effective method WEIN for remote sensing image ocean front detection. our WEIN model performs weak edge identification by CNNs, and obtains the ocean front edge image of each stage through the side output layer, and finally fuses to attain accurate ocean front detection results. The model is optimized using the binary cross entropy objective function and the IoU objective function for the ocean front detection task. The experimental results show the effectiveness of WEIN.\par
In our work, the production of the ocean front data set is also a task that cannot be ignored. Due to the immaturity of the previous ocean front detection technology, there has been no relevant ocean front data set for researchers to use. In order to promote the development of ocean front detection, we collect satellite remote sensing images and process it to obtain ocean front gradient images, and ask relevant experts to analyze and calibrate to obtain groundtruth. In the future, we may disclose the ocean front data set for later research.

\ifCLASSOPTIONcaptionsoff
  \newpage
\fi

\end{document}